\documentclass{article}

\PassOptionsToPackage{numbers, compress}{natbib}

\usepackage[preprint]{neurips_2021}

\usepackage[compact]{titlesec}
\usepackage[utf8]{inputenc} 
\usepackage[T1]{fontenc}    
\usepackage{hyperref}       
\usepackage{url}            
\usepackage{booktabs}       
\usepackage{amsfonts}       
\usepackage{nicefrac}       
\usepackage{microtype}      
\usepackage{xcolor}         

\usepackage{graphicx}
\usepackage{adjustbox}
\usepackage{array}
\usepackage{subcaption}
\usepackage{multirow}
\usepackage{soul}
\usepackage{courier}
\usepackage{enumitem}
\usepackage{amsmath}
\usepackage{amssymb}
\usepackage{algorithmicx}
\usepackage{algorithm,algpseudocode}
\algnewcommand{\algorithmicforeach}{\textbf{for each}}
\algdef{SE}[FOR]{ForEach}{EndForEach}[1]
  {\algorithmicforeach\ #1\ \algorithmicdo}
  {\algorithmicend\ \algorithmicforeach}
\usepackage{csquotes}
\DeclareMathOperator*{\argmax}{arg\,max}

\graphicspath{{./figs/}}
\hypersetup{
    colorlinks=true,
    linkcolor=red,
    citecolor=blue,
}

\title{Building Object-based Causal Programs for Human-like Generalization}

\author{%
  Bonan Zhao \\
  Department of Psychology \\
  University of Edinburgh \\
  \texttt{b.zhao@ed.ac.uk} \\
  \And
  Christopher G. Lucas \\
  School of Informatics \\
  University of Edinburgh \\
  \texttt{clucas2@inf.ed.ac.uk} \\
  \AND
  Neil R. Bramley \\
  Department of Psychology \\
  University of Edinburgh \\
  \texttt{neil.bramley@ed.ac.uk} \\
}

\begin{document}

\maketitle

\begin{abstract}
  We present a novel task that measures how people generalize objects' causal powers based on observing a single (Experiment 1) or a few (Experiment 2) causal interactions between object pairs.
  We propose a computational modeling framework that can synthesize human-like generalization patterns in our task setting, and sheds light on how people may navigate the compositional space of possible causal functions and categories efficiently.
  Our modeling framework combines a causal function generator that makes use of agent and recipient objects' features and relations,
  and a Bayesian non-parametric inference process to govern the degree of similarity-based generalization.
  Our model has a natural “resource-rational” variant that outperforms a na{\"i}ve Bayesian account in describing participants, in particular reproducing a generalization-order effect and causal asymmetry observed in our behavioral experiments.
  We argue that this modeling framework provides a computationally plausible mechanism for real world causal generalization.
\end{abstract}

\section{Introduction}

Objects appear to be a fundamental building block of our world models, appearing early in development and ubiquitously in natural language \citep{baillargeon1995physical,spelke1990principles,spelke2007core}. 
Observing objects interacting naturally invokes causal perceptions.
For instance, in ``launching'' phenomena \cite{michotte1963perception},
when participants observe some object A moving toward a stationary object B, and if around when A touches B, A stops moving and B starts to move, participants spontaneously report that they see object A cause object B to move \citep[see also][]{gordon1990perceived,leslie1987six,scholl2000perceptual}.

While a wealth of research has been devoted to studying how children and adults acquire causal beliefs \cite[e.g.,][]{sloman2005models,bramley2015conservative,gopnik2007causal,griffiths2009theory,kemp2012integrated}
and generalize functional properties \cite[e.g.,][]{goodman2008rational,lake2015human,shepard1987toward,tenenbaum2001generalization,wu2018generalization},
the interplay between causality, object concepts and generalization has received less attention.
On the face of it, this is surprising.
In reality, a key component of successful causal learning is the ability to generalize causal relations appropriately to new situations that are related but non-identical to past experiences \cite{pearl2011transportability,bareinboim2013general,bareinboim2016causal}.
Meanwhile, generalization could not be successful without tapping into  what Sloman calls Nature's ``invariants'', the true causal laws that govern both experienced and novel situations \cite{sloman2005models}.
Recent research has explored this interplay using hierarchical Bayesian models \cite[e.g.,][]{griffiths2009theory,kemp2010learning,goodman2011learning} as a computational level account of domain knowledge \citep{marr1982vision},
or formal analysis on transportability of structual causal knowledge \cite{pearl2011transportability,bareinboim2013general,bareinboim2016causal},
but these are limited in their ability to capture psychological processes due to their inherent intractability \citep{kwisthout2020computational,van2008tractable,valentin2021symbolic}.

\begin{figure*}[t]
	\centering
  \begin{subfigure}[t]{0.148\textwidth}
  	\centering
  	\includegraphics[width=\linewidth]{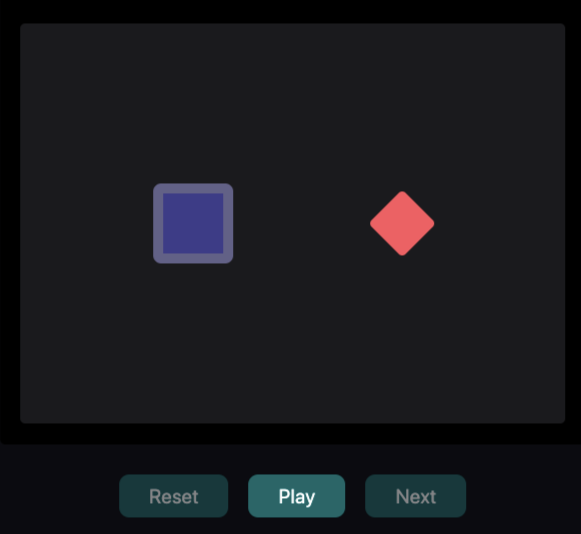}
  	\caption{Start}
  \end{subfigure}
  \hfill
  \begin{subfigure}[t]{0.148\textwidth}
  	\centering
  	\includegraphics[width=\linewidth]{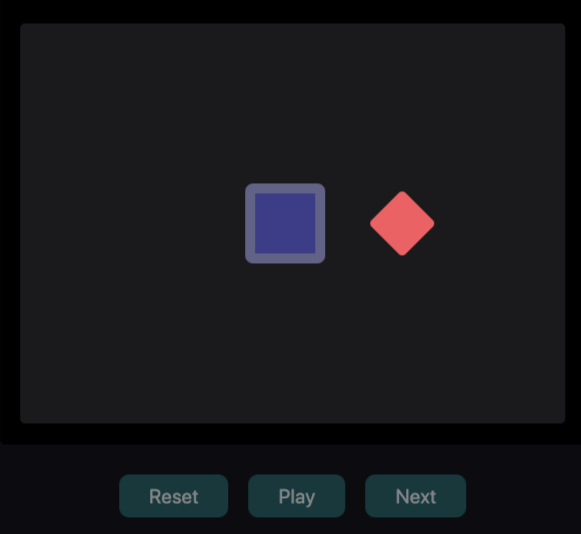}
  	\caption{Animation}
  \end{subfigure}
  \hfill
  \begin{subfigure}[t]{0.148\textwidth}
  	\centering
  	\includegraphics[width=\linewidth]{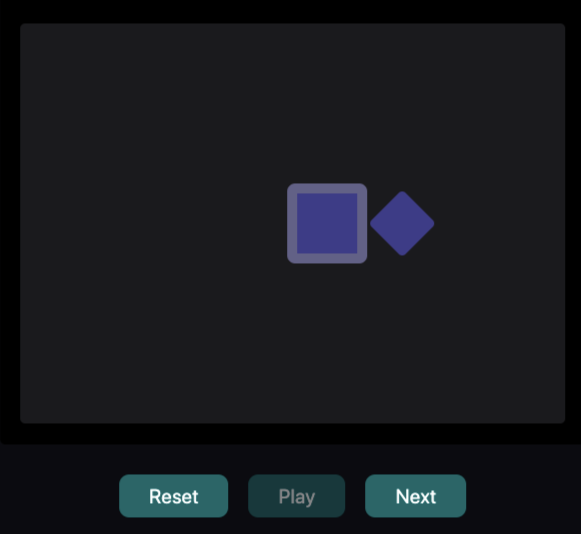}
  	\caption{Effect}
  \end{subfigure}
  \hfill
  \begin{subfigure}[t]{.49\textwidth}
  	\centering
  	\includegraphics[width=\linewidth]{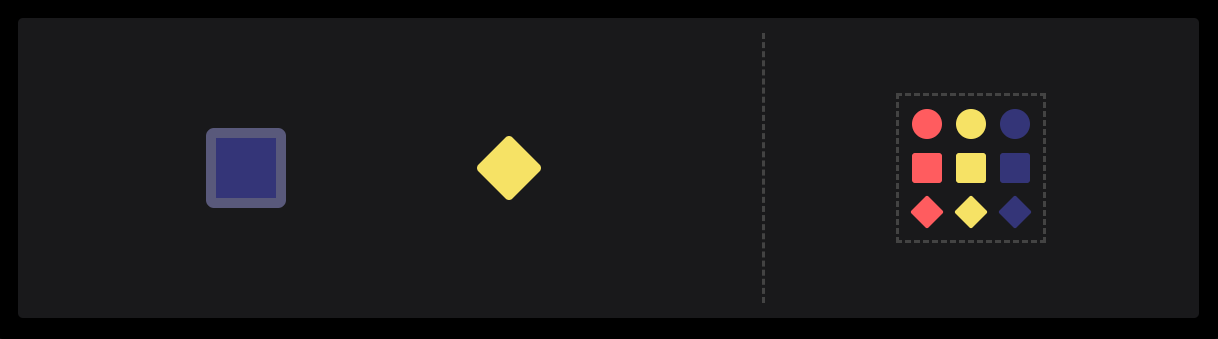}
  	\caption{Generalization task (selection panel)}
  \end{subfigure}

  \vspace{1em}
  \begin{subfigure}[t]{0.49\textwidth}
    \centering
    \fcolorbox{black}{white}{\includegraphics[width=\linewidth]{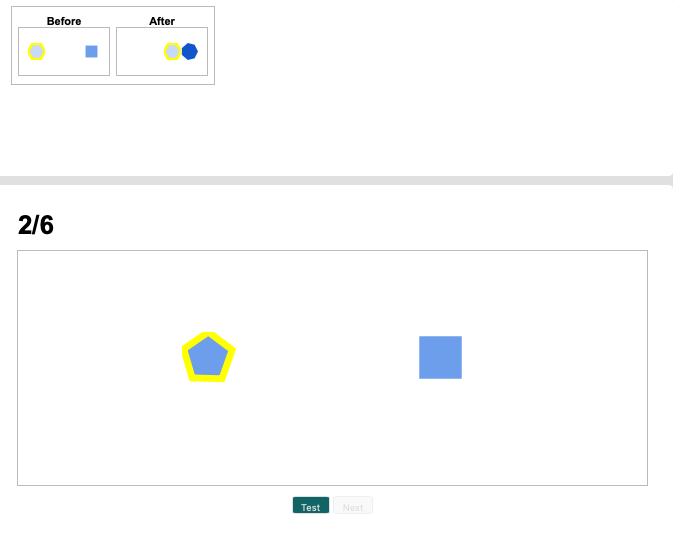}}
    \caption{Learning example}
  \end{subfigure}
  \hfill
  \begin{subfigure}[t]{0.49\textwidth}
    \centering
    \fcolorbox{black}{white}{\includegraphics[width=\linewidth]{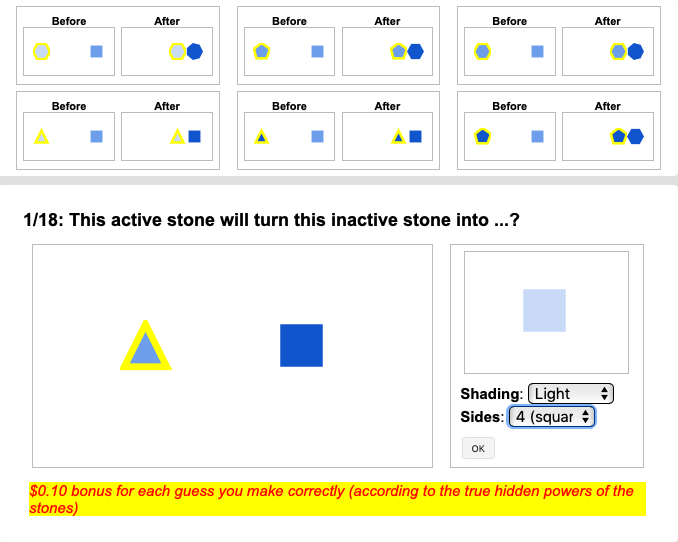}}
    \caption{Generalization task (drop-down menu)}
  \end{subfigure}

\caption{
Task interfaces.
  A--D: Experiment 1. A--C step through an example learning scene animation, and
  D shows a generalization task consisting of novel objects (left) and a selection panel (right), in which learners select from a set of possible predictions about the appearance of the recipient after the causal interaction.  
  E--F: Experiment 2. Summaries of previous learning examples are shown at the top of the screen.
  E shows one animated effect similar to A--C.
  In F, generalization predictions are elicited by selecting from two drop-down menus (one per feature).
  }
\label{fig:exp_interface}
\end{figure*}

In this paper, we explore how people generalize causal relations from observed interactions between pairs of simple geometric objects, and propose a computational modeling framework that has a natural “resource-rational” variant.
We develop an interactive online game we call ``magic stones'', in which people can test how one object (the agent) acts upon another (the recipient) and brings about some change in the recipient (see Figure~\ref{fig:exp_interface}A--C),
and then make predictions about new pairs of objects (Figure~\ref{fig:exp_interface}D).
This game thus provides behavioral measures of how people {\em generalize} their causal understanding of observed objects to unseen ones.

In the following sections, we introduce our computational modeling framework for object-based causal inference with an expressive hypothesis space that captures the diverse inferences people can make \cite{griffiths2009theory,lucas2010learning}.
It draws on non-parametric approaches to category and function learning to account for similarity-based generalization predictions (Figure~\ref{fig:model-sum}).
The normative version, we call LoCaLa (Local Causal Laws), compares each generalization trial against all the learning examples in order to assign causal categories to new observations.
We then describe a ``resource-rational"  \citep{goodman2008rational,sanborn2010rational} variant, we call LoCaLaPro (Local Causal Laws Process), that shares causal categories among generalization trials, and only posits a new causal function and category when it cannot explain a novel observation with any existing categories.

We report on two experiments that shed light on previously unexplored inductive biases in causal learning, and so allow us to evaluate our models and the ideas that motivate them.
We find that our local laws and particularly our new process model better explain our behavioral data than a purely normative account, including explaining a novel generalization-order effects observed in Experiment 1,
and causal asymmetry in Experiment 2.

\renewcommand\thesubfigure{\Alph{subfigure}}

\section{Related work}

Our task generalizes the structure of standard ``blicket detector" studies, in which different combinations of factors or objects are tested and an effect does or does not occur \citep[e.g.,][]{gopnik2007causal,kemp2010learning,lucas2010learning,sim2017learning}.
The collision stimuli we used in our tasks are known to evoke automatic perception of causality \citep{michotte1963perception},
making these an appealing way to study how people inherently reason about cause and effect.
In daily life, we typically observe sequences of changes rather than independent trials
\citep{johnson2015causal,soo2018causal,steyvers2003inferring}, and our experiment interface can capture this pattern naturally.
Unlike previous work \citep[e.g.][]{kemp2010learning,lucas2014children}, we are not constrained to binary, present/absent effects, or multiple outcomes, such as different kinds of activations \citep[e.g.][]{schulz2006god}.
Our task can also capture higher-order causal relationships, such as rules that depend on color/shape matches between agent and recipient objects \citep{sim2017learning}.

Causal Bayes nets (CBN) \cite{pearl2000causality,pearl2009causality} have inspired fruitful research in causal learning and induction,
among which the most relevant to causal generalization are transportability analysis \cite{pearl2011transportability,bareinboim2013general,bareinboim2016causal} and
the hierarchical Bayesian model (HBM) framework \cite{griffiths2009theory,kemp2010learning,lucas2010learning}.
However, these methods have been focused on inferring statistical relationships between variables, rather than interactions between objects.
Furthermore,
although CBNs and HBMs
can address the problem of inferring network structure together with parameter estimation, they suffer from serious scalability issues:
As the number of nodes and layers increases, the number of possible causal networks increases super exponentially, making domain expectations more of a computational curse than a learning-to-learn blessing in practice \cite{kwisthout2020computational,van2008tractable}.
Therefore, more recent accounts of causal learning have treated causal inference as practically constituting a search problem in a large multi-modal theory space \citep{griffiths2009theory,kemp2012integrated,goodman2008rational,kemp2010learning,goodman2011learning,bramley2017formalizing},

While it has been argued that we think of causal relationships as ``invariant'' \citep{sloman2005models}, category knowledge is also integral to real world causal inference.
While people refer to causal relationships when categorizing objects \citep{gopnik2000detecting,rehder2001causal,rehder2003categorization},
they also spontaneously use featural and relational information for categorization when no causal information is available to form categories \citep[][]{anderson1991adaptive,love2004sustain,Kemp2008discovery},
and then make causal predictions based on these categories \citep{kemp2010learning}.
Since creation of such causal categories may be triggered only when required for generalization, we will present a process view such that cognitive representations are fundamentally generative, and judgments are based on samples \citep{bramley2017formalizing,chater2018mind, stewart2006decision}.

\section{Formulation}

\begin{figure}
  \centering
  \includegraphics[width=.9\textwidth]{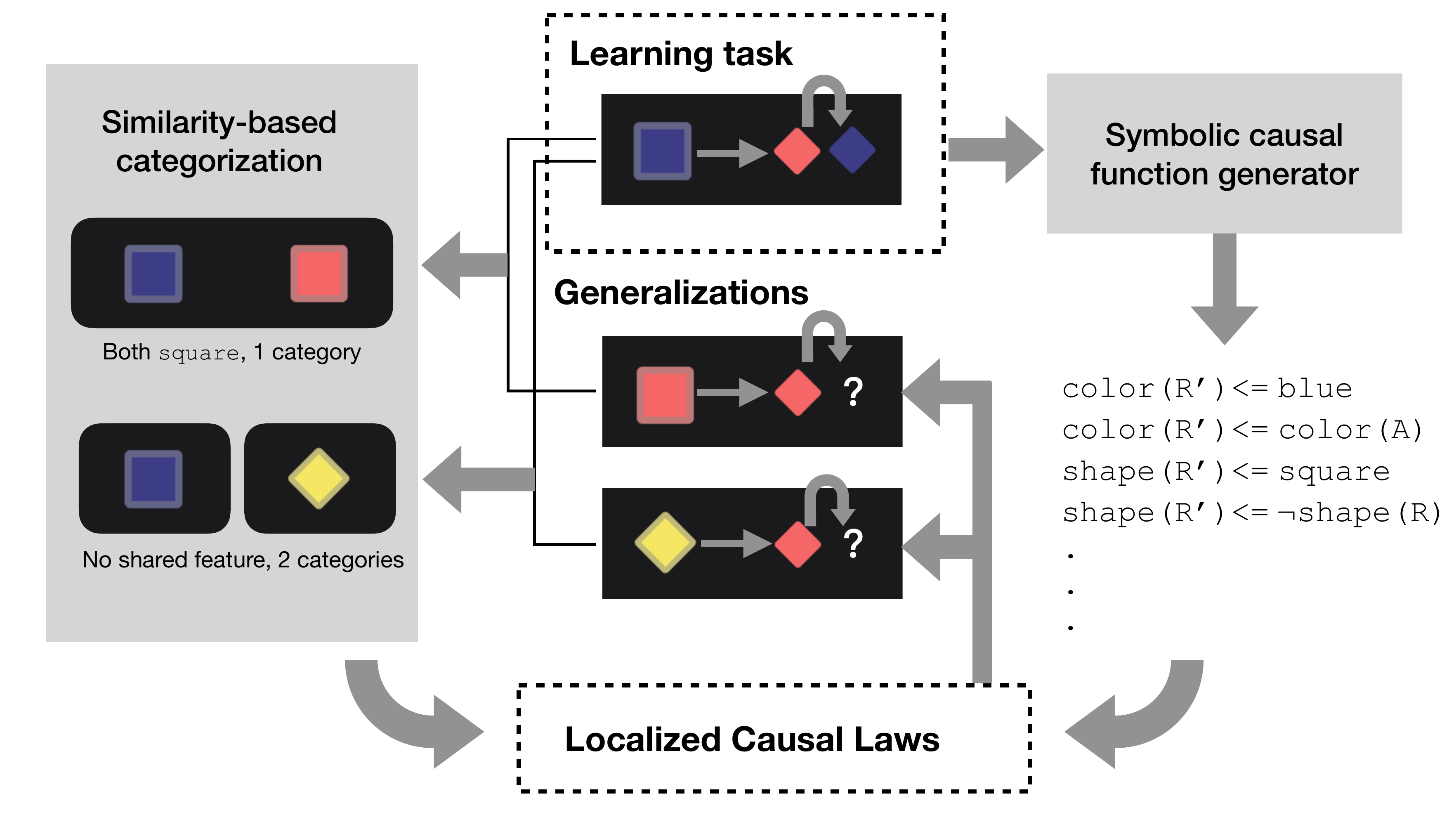}
  \caption{To model how people make object-based causal generalization predictions (middle),
  we combine program induction about the hidden causal laws (right) with non-parametric category inference about their domains of influence (left).
  Together, they form causal categories that guide generalization predictions (arrows from bottom to middle).}
  \label{fig:model-sum}
\end{figure}

\paragraph*{Causal functions} We use a Probabilistic Context-Free Grammar \citep[PCFG;][]{ginsburg1966mathematical} to define a prior over possible causal functional relationships (causal laws).
Grammar $\mathcal{G}$ generates expressions that specify features of the result object.
For example, if one of the features is ``color'', a possible causal function could be
$\texttt{color}(r') \Leftarrow \texttt{red}$
 --- recipient will turn red --- or
$\texttt{color}(r') \Leftarrow \texttt{color}(a)$
--- recipient will take the agent's color, and so on.
This grammar is set up to allow for arbitrarily complex expressions
allowing a rule to produce conjunctions of feature changes,
for example,
$\textsc{and}(\texttt{color}(r') \Leftarrow \texttt{red}, \texttt{shape}(r') \Leftarrow \texttt{triangle})$.
A causal function outputs result object(s) when particular agent and recipient objects are provided.
Detailed definitions for our grammar $\mathcal{G}$ are provided in Appendix~\ref{ap:pcfg}.

\paragraph*{Latent causal categories}
While this PCFG provides an account for the set of possible causal functions that people may entertain during learning (right box in Figure~\ref{fig:model-sum}), it cannot tell us to what extent should these causal laws apply. 
When making generalization predictions, if the novel objects look similar to those in the learning phase, one may consider these objects falling into one category and are thus governed by the same causal law.
However, if the novel objects look very different from learning examples, these objects may belong to different categories and therefore execute different causal functions (left box in Figure~\ref{fig:model-sum}).

We formalize the idea that the objects may fall into different causal categories with respect to featural similarities, roles in the interaction, and shared causal laws.
Let $\mathbf{d}$ denote a set of observations,
$\mathbf{z}$ denote a particular set of causal category memberships,
and $\mathbf{w}$ some categorization parameters (weights).
We use superscript $(i)$ for the $i$-th observation:
$d^{(i)}$ for the $i$-th data point,
$z^{(i)}$ the causal category assigned to the $i$-th observation,
$a^{(i)}$ the agent in the $i$-th data point, similarly for $r^{(i)}, r'^{(i)}$,
and $\mathbf{w}_{z^{(i)}}$ for the weights associated with categorizy $z^{(i)}$;
additionally, let $z^{(-i)}$ be the categorization of observations except for $d^{(i)}$,
inference about the $i$-th observation's category is given by:
\begin{align}
  P(z^{(i)}|\mathbf{d}, \mathbf{w})
    &= P(z^{(i)}|d^{(i)}, \mathbf{w}, z^{(-i)}) \nonumber \\
    &\propto P(z^{(i)}|z^{(-i)}) P(a^{(i)}, r^{(i)}|\mathbf{w}_{z^{(i)}}) P(r'^{(i)}| a^{(i)}, r^{(i)}, \mathbf{w}_{z^{(i)}})
   \label{equ:cat}
\end{align}
Equation~\ref{equ:cat} consists of three parts:
$P(z^{(i)}|z^{(-i)})$ reflects our expectations about how causal categories are distributed,
$P(a^{(i)}, r^{(i)}|\mathbf{w}_{z^{(i)}})$ encodes our beliefs about object features and category membership,
and $P(r'^{(i)}|a^{(i)}, r^{(i)}, \mathbf{w}_{z^{(i)}})$ marks the causal function this particular category possesses.
To accommodate the fact that there could be any number of causal categories,
we draw on an extended Dirichlet Process,
composed by a Chinese Restaurant Process (CRP) \citep{aldous1985exchangeability}, a multinomial distribution over the feature values of an unknown number of categories with a Dirichlet prior,
and likelihoods defined by causal functions,
in correspondence to the three parts in Equation~\ref{equ:cat}.
Appendix~\ref{ap:model} unpacks this procedure
in more details.

In total, we introduce three global parameters for our extended Dirichlet Process:
a concentration parameter $\alpha > 0$ for the distribution of categories according to CRP,
a Dirichlet prior $\beta \ge 0$ to control the impact of feature similarities,
and a focus parameter $\gamma\in [0, 1]$ for weighting the categorization based on causal action roles (agent or recipient).
In our Dirichlet Process, both the focus parameter $\gamma$ and the Dirichlet prior $\beta$ are embedded in a local parameter $\mu^{(z_i)}$, the mean feature vector for category $z^{(i)}$ (see Appendix~\ref{ap:model}).
Let $f^{(z_i)}$ be the causal function assigned to category $z^{(i)}$,
we can rewrite Equation~\ref{equ:cat} as:
\begin{equation}\label{equ:dp_core}
  P(z^{(i)}|\mathbf{d},\mathbf{w}) \propto P(z^{(i)}|z^{(-i)}) P(a^{(i)}, r^{(i)}| \mu^{(z_i)}) P(d^{(i)}|f^{(z_i)} )
\end{equation}

\paragraph*{Inference} The Dirichlet Process defined by Equation~\ref{equ:dp_core} models the acquisition of causal categories,
equivalent to the learning phase of causal generalization.
It is impossible to compute the posterior directly because we do not know how many categories are there in advance,
but we can approximate the posterior distribution using Gibbs sampling (see also Appendix~\ref{ap:model}).
In the prediction stage, upon observing a partial data point $d^* = (a^*, r^*, \cdot)$, an optimal decision can be made by marginalizing over the posterior predictive distribution of each possible $r'^{*}$ value:
\begin{align}
  P(\tilde{d^*}) &\propto \int_{z} p(\tilde{d^*}|z)P(z|d)\text{d}z \nonumber \\
  &\approx \frac{1}{|\tilde{Z}|} \sum_{\tilde{z} \in \tilde{Z}} p(r'^*|a^*, r^*, f^{(\tilde{z})})P(a^*, r^*|\mu^{(z)})P(z|d)
  \label{eq:predictive_full}
\end{align}
and taking the maximum over this predictive posterior:
\begin{equation}
\mathrm{Choice} = \argmax P(\tilde{d^*})
\label{eq:hard_max}
\end{equation}

\paragraph*{Process variant} Instead of trying to approximate a global optimal distribution of latent causal categories,
we further develop a process model that commits to its own causal category allocations as it makes generalizations,
and assigns new observations to a new or existing category
according to category sizes and objects' featural similarity:
\begin{equation}\label{eq:feat_cat}
  P(z^{(i)}|a^{(i)}, r^{(i)}) \propto P(z_i|z^{(-i)})P(a^{(i)}, r^{(i)}|\mu^{(z_i)}).
\end{equation}

Equation~\ref{eq:feat_cat} is a simplified version of Equation~\ref{equ:dp_core}, because in generalization scenarios, the resulting $r'$ is unknown and the assignment is purely based on the basis of feature-based fit to existing categories.
Instead of approximating a posterior over infinitely many possible categories,
this process model maintains a small set of available categories that are created online as new generalizations are performed.
Interested readers can find more implementation details in Appendix~\ref{ap:process}.

\section{Evaluations}

\subsection{One-shot causal generalization}

\paragraph*{Data} We designed a one-shot causal generalization experiment (Figure~\ref{fig:exp_interface}A--D) involving six different one-shot learning tasks (see Appendix~\ref{ap:exp1-setup}).
We collected data from one-hundred-and-twenty participants (53 female, aged 40$\pm$11) from Amazon Mechanical Turk.
Each participant faced a single learning task and made 15 generalization predictions, leading to 1800 generalizations in total.

In addition, we distinguished two generalization sequences for each one-shot learning task: a {\em near-first transfer} and a {\em far-first transfer}.
In the near-first transfer condition, generalizations start with cases
that differ by only one feature from the learning example and progress to cases in which all of the features differ.
In the far-first transfer condition,
generalizations are first made about sets of objects that are completely different from those in the learning examples and progress back to the more similar cases.

\begin{figure*}[t]
	\centering
  \includegraphics[width=\linewidth]{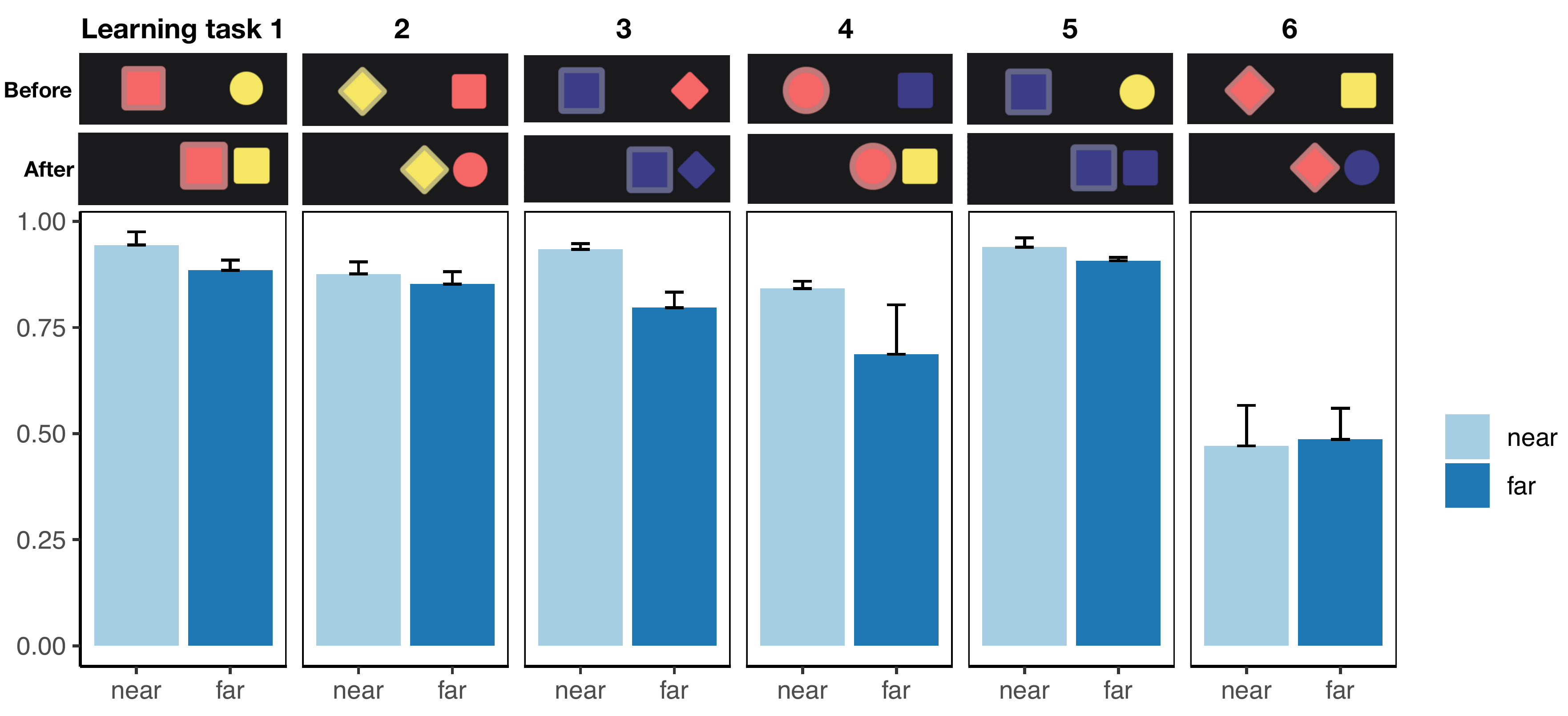}
  \caption{Experiment 1 generalization consistency $\rho_\tau$ (y-axis), averaged over generalizations per each one-shot learning task (as illustrated on top of each panel) and sequence order: light blue = \emph{near-first transfer}, dark blue = \emph{far-first transfer}.}
  \label{fig:exp1-bars}
\end{figure*}

\paragraph*{Behavioral results} Since there is no ground truth in this task to measure accuracy against, we use Cronbach's alpha \citep{cronbach1943estimates} to measure inter-participant generalization consistency $\rho_\tau$ for each generalization under each learning task and order condition. Since our design is completely between-subject, this tests how peaked the distribution of responses was on average across the sample of participants in that condition and task (e.g. each corresponding to the rows in Figure~\ref{fig:exp1-bars}A).
Fisher's exact test confirms that participants' generalization consistency ($\rho_\tau =.80\pm.22$) is significantly above chance, $p<0.001$,
demonstrating a robust human capacity to make systematic one-shot causal generalizations \citep{kemp2007learning}.

Near-first transfers induced more consistent predictions across subjects ($\rho_\tau = .83\pm.21$),
compared with far-first transfers ($\rho_\tau = .77\pm.21$),
$t(89)=3.54, p<.001$, 95\% CI = [0.03, 0.10].
Inter-person generalization consistency $\rho_\tau$ was higher for near-first transfer under all learning conditions except A6 \emph{``Recipient changes to a new color and shape''}, for which both transfer sequences induced low agreement (Figure~\ref{fig:exp1-bars}).
This generalization-order effect suggests that participants may be influenced by their own generalization history in some way.

\begin{figure}[t]
  \centering
  \includegraphics[width=\textwidth]{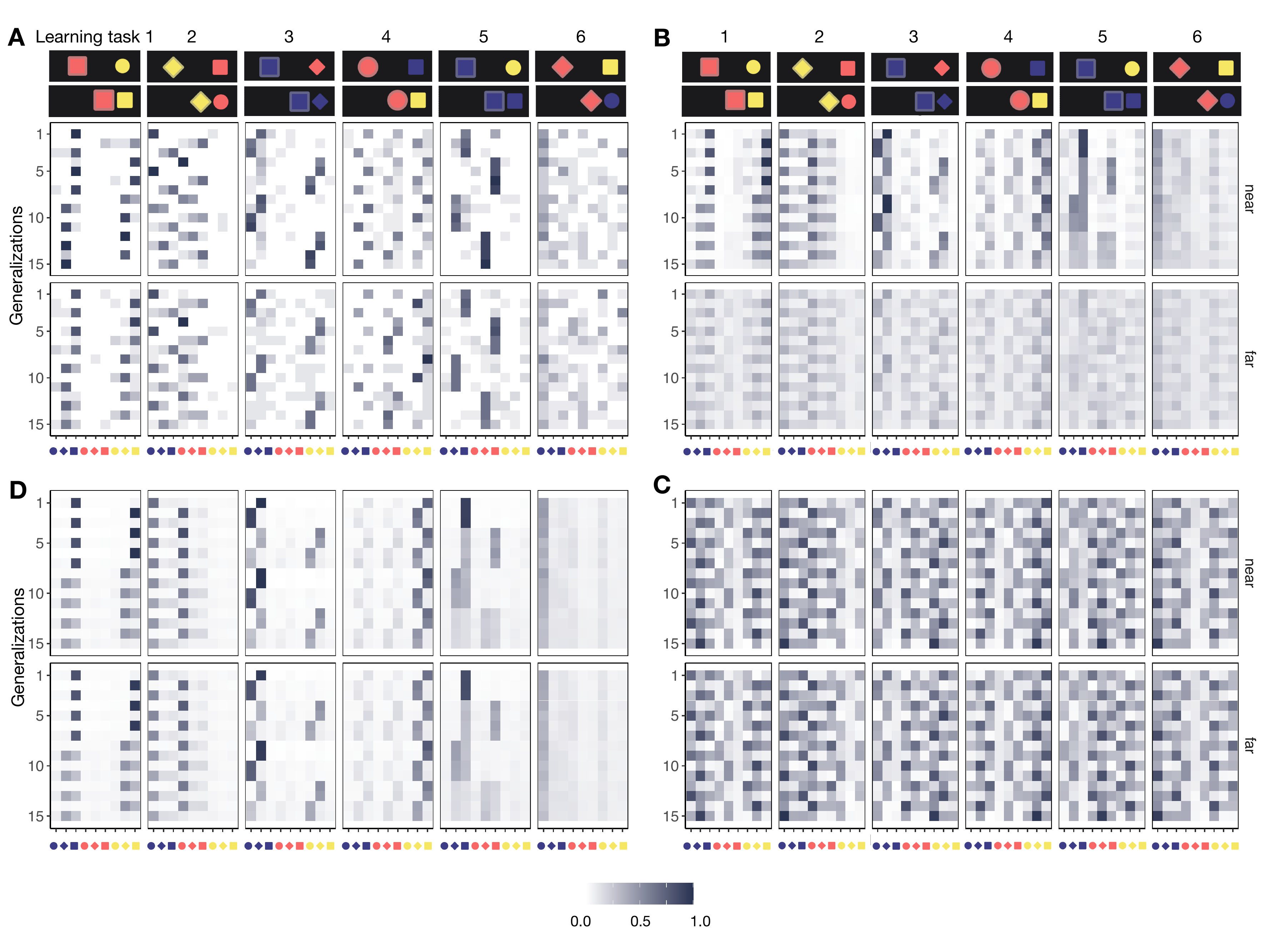}
  \caption{Experiment 1. A. Generalization consistency patterns for all conditions visualized as proportion of participants predicting each stone type for $r'$ (column) on each task (row).
  B-C. Example LoCaLaPro predicted proportions with small $\alpha$ (=0.01) and large  $\alpha$ (=8). For both figures, $\beta=0, \gamma=0.5$.
  D. Fitted LoCaLaPro predictions.
  }
  \label{fig:exp1-mod}
\end{figure}

\paragraph*{Models} We fit several model variants to our choice data (Figure~\ref{fig:exp1-mod}A) using maximum likelihood, and then compared them using Bayesian Information Criterion to accommodate for different numbers of parameters.
The random choice \emph{Baseline} model simply predicts $P(\mathrm{choice}=r') = \nicefrac{1}{9}$, for the 9 candidate objects and has no parameters.
The \emph{Universal Causal Laws (UnCaLa)} model uses the PCFG-generated causal functions,
assuming that the causal function governing the training case applies universally to all potential generalization scenarios, no matter how dissimilar the objects involved may be.
The \emph{Local Causal Laws (LoCaLa)} model implements the joint inference of latent causal categories, combining both symbolic causal laws and similarity-based categorizations.
Finally, the \emph{Local Causal Laws Process (LoCaLaPro)} model is the averaged predictions of our process version of the LoCaLa model.

For the two causal category models LoCaLa and LoCaLaPro, we fit hyper parameters $\alpha$ and $\beta$, but fixed the focus parameter $\gamma = 0.5$ because there is no information about what causal categorization assumptions should be preferred in this experiment. 
For all of the non-random models, we applied a softmax with a ``inverse temperature'' parameter $t$ on the posterior predictives  to account for response noise \citep{luce1959individual}.

\paragraph*{Model fits} Table~\ref{tab:model-result} summarizes model fits.
Both the UnCaLa and LoCaLa models improve dramatically over the random Baseline,
and the LoCaLa outperforms the UnCaLa in both log likelihood and BIC.
The process model LoCaLaPro best predicts the empirical data, and as shown in
Figure~\ref{fig:exp1-mod}D, it indeed predicts the dominant judgment patterns among participants.

Figure~\ref{fig:exp1-mod}B-C demonstrate that when concentration parameter $\alpha$ is small, it can reproduce a strong generalization-order effects, because each new observation will stickily join previously assigned categories (Figure~\ref{fig:exp1-mod}B).
When $\alpha$ becomes very large, however, a new observation has a high probability of being attributed to a new category (Equation~\ref{eq:crp}), and the overall generalization predictions will simply approach the prior (Figure~\ref{fig:exp1-mod}C).
The fitted $\alpha$ parameter for LoCaLaPro is $0.38$, confirming the presence of a dominant order effect.

\subsection{Multi-shot causal generalization}

\paragraph*{Data} We extended the setup in Experiment 1 to investigate inference from multiple complete observations, where each participant tested six pairs of objects before making generalizations.
We controlled whether participants observe the same agent object paired with various recipient objects (fixed-agent conditions), or {\em vice versa} (fixed-recipient conditions).
We employed two ground truth rules between-subject to counterbalance between shape and shading features (see Appendix~\ref{ap:exp2-setup}).
One-hundred-and-sixty-three participants were recruited from Amazon Mechanical Turk.
Sixty-one participants were excluded before analysis for failure to provide task-relevant responses, 
resulting in 102 participants (37 female, aged $35 \pm 10$) $\times$ 16 generalization predictions $= 1632$ generalizations.

\paragraph*{Behavioral results} As with Experiment 1, we measured inter-person generalization consistency $\rho_T$, and Fisher's exact test confirmed that
across all experimental conditions, participants produced systematic generalization patterns against random guesses, $p < 0.001$.
In particular, the {\em fixed-agent} condition induced higher consistency ($\rho_T = 0.89 \pm 0.06$) than the {\em fixed-recipient} condition ($\rho_T = 0.85 \pm 0.1$), $t(31) = 2.12, p = 0.04, 95\%\text{CI} = [0.001, 0.08]$,
while the difference in $\rho_T$ between the ground truth condition was negligible, $t(31) = 0.22, p=\text{n.s.}$ (see also Appendix~\ref{ap:exp2-gen}-\ref{ap:exp2-selfr}).
This suggests that participants made more homogeneous predictions after observing the same agent acting on a range of recipients,
and diverged more having observed different agents interacting on the same recipient,
echoing a well-known inductive bias---causal asymmetry---in physical causation \citep{white2006causal}.

\paragraph*{Models} As with Experiment 1, we compared participants generalizations to a random \emph{Baseline} model,
a \emph{Universal Causal Laws (UnCaLa)} and a \emph{Local Causal Laws (LoCaLa)} model, again using maximum likelihood and BIC to account for different numbers of parameters.
Since we randomized the presentation of both evidence and generalization trials between subjects, we do not expect systematic effects of the sort accommodated by our process model LoCaLaPro, so focus on comparison between UnCaLa and LoCaLa. 
Different from Experiment 1, values of $\gamma = 1, 0.5$ and $0$ are of particular theoretical interest here,
representing categorization based on just the agent, agent and recipient equally, or just the recipient. We also included $\gamma = 0.25$ and $\gamma = 0.75$ consistent with a mixed focus biased toward either agent or recipient.

\begin{table}[t]
  \caption{Model fitting results}
  \label{tab:model-result}
  \begin{tabular*}{\textwidth}{@{\extracolsep{\fill}} lllllll}
    \toprule
              & $\alpha$ & $\beta$ & $\gamma$ & $t$     & Log likelihood & BIC         \\
    \midrule
    \multicolumn{7}{c}{Experiment 1} \\
    Baseline  &          &         &          &         & -3955          & 7910        \\
    UnCaLa    &          &         &          & 6.96    & -2761          & 5529        \\
    LoCaLa    & 2.41     & 938.81  & (0.5)    & 9.44    & -2748          & 5518        \\
    LoCaLaPro & 0.38     & 1       & (0.5)    & 10.09   & {\bf -2736}    & {\bf 5494}  \\
    \midrule
    \multicolumn{7}{c}{Experiment 2} \\
    Baseline  &          &         &          &        & -4889           & 9778        \\
    UnCala    &          &         & (0.5)    & 3.19   & -3706           & 7417        \\
    LoCaLa    & 9        & 256     & 1        & 9.5    & {\bf -3462}     & {\bf 6942}  \\
    \bottomrule
  \end{tabular*}
\end{table}

\paragraph*{Model fits} As summarized in Table~\ref{tab:model-result}, both models improve substantially over the random Baseline, with LoCaLa fitting better than UnCaLa as in Experiment 1.
Within LoCaLa, the best fitting $\gamma$ value was 1, indicating that causal categorization was dominated by features of the agents,
in line with the asymmetric causal attribution bias suggested by our regression analyses.
The fitted $\alpha$ for LoCaLa is 9, above chance-level probability of assigning a new causal law to each new observation, confirming the behavioral tendency to create multiple causal categories to account for the evidence.
Here, $\gamma = 1$ together with $\alpha = 9$ captures the causal asymmetry observed in behavioral data: When observing multiple different agents, participants imputed many local causal laws. When seeing a single agent interact with multiple recipients, they tended to impute a single causal law.
The fitted $\beta$ parameter was quite large, as in Experiment 1, indicating a substantial heterogeneity across participant data taken together.

\section{Discussion}

In this paper, we investigated causal generalization based on observations of interactions between objects.
Our two experiments demonstrated that people make systematic causal generalizations from one or a few observations and revealed some of the inductive biases that drive these.
Participants' generalization patterns were well-captured by our Bayesian inference model
operating on a latent space of causal laws generated by a simple Probabilistic Context Free Grammar prior favoring parsimony, 
and an extended Dirichlet Process that localized causal laws according to the interacting objects' features as well as their causal behaviors. Separately, these ideas extend previous work in causal inference and categorization \citep{goodman2008rational,kemp2010learning,bramley2017formalizing}, and in combination they give the first precise formal account of how
people (1) partition the world according to causal behavior without relying on innate knowledge -- an essential feature of any general model of causal learning \citep[e.g.][]{griffiths2009theory,lucas2010learning};
and (2) do so in a way that is resource-efficient, requiring modest attention and memory, and supporting snap judgments, albeit at the expense of inducing order effects.

Our framework integrates a symbolic approach to represent causal law generation,
with non-parametric Bayesian categorization to model latent categories, emphasizing the constructive nature of causal belief formation, in which both the content and extension of our causal concepts are generated rather than pre-specified.
The constructive nature of the PCFG calls upon a potentially infinite set of possible causal functions, yet is governed by the preference for parsimony, and encourages systematic composition \citep[see also][]{goodman2008rational,bramley2018grounding}.
The extended Dirichlet Process for category construction goes beyond a hierarchical Baysian modeling approach where categories are pre-defined as inductive biases \citep[e.g.][]{griffiths2009theory,goodman2011learning}, and thus better captures the flexibility of human generalization behaviors \citep[see also][]{kemp2010learning}.
This method draws a close link with probabilistic program induction models \citep[e.g.][]{lake2015human,bramley2018grounding,ellis2021dreamcoder,lake2020people}, where causal beliefs and concepts can be viewed as programs, and accurate generalizations can be viewed as evidence for successful program synthesis whereby these programs increasingly reflect the true causal laws of nature.
Moreover, our constructive computational modeling framework balances between learning a single causal law versus making generalization predictions based on multiple causal categories,
and with the ``creating new categories only on demand'' assumption for a process account, our model successfully reproduces the generalization-order effects in behavioral data.

However, our simple task and specific modeling choices just mark the starting point for exploring causal generalization.
Our modeling framework is compatible with many other options.
For example, one may to extend the symbolic approach to cover the categorization process as well,
or use causal Bayes net as an alternative representation for causal functions.
Our task setup can be adapted to investigate causal interactions between multiple objects, reversed agent-object roles, or active learning and planning tasks.
In our current work-in-progress, we are extending the existing framework to use adaptor grammar \citep{liang2010learning,o2009fragment,briggs2006functional} to model organic compositional causal generalization under boostrapping conditions.
We hope this line of work opens up more insights into how causality permeates the cognitive representations we use to predict, explain, and act in the world.

\bibliographystyle{unsrt}
\bibliography{bib}

\clearpage
\pagenumbering{arabic}
\renewcommand*{\thepage}{A\arabic{page}}

\begin{changemargin}{0.7cm}{0.7cm}
\centering
{\Large Supplementary material for paper "Building Object-based Causal Programs for Human-like Generalization"}
\end{changemargin}

\vspace{0.5cm}

\appendix

\section{PCFG for symbolic causal functions}\label{ap:pcfg}

Let $f$ be a causal function that takes the agent object $a$ and the recipient object $r$ as input,
and outputs the result object $r'$.
We assume a prior over possible causal functions in the form of a PCFG $\mathcal{G} = (\Gamma,\Theta)$, 
where $\Gamma$ is a set of production rules and $\Theta$ is a set of production 
probabilities (see Table~\ref{tab:core_grammar}).
Let $\phi_i$ denote the $i$-th feature in the set of all observable object features $\Phi$.
Grammar $\mathcal{G}$ thus generates expressions that specify features of the result object.
For example, if one of the features is ``color'', a possible causal function could be
$\texttt{color}(r') \Leftarrow \texttt{red}$
 --- recipient will turn red --- or
$\texttt{color}(r') \Leftarrow \texttt{color}(a)$
--- recipient will take the agent's color, and so on.
The grammar is set up to allow for arbitrarily complex expressions through the the ``bind additional'' production rule (Table~\ref{tab:core_grammar}, row 2), allowing a rule to produce conjunctions of feature changes,
for example,
$\textsc{and}(\texttt{color}(r') \Leftarrow \texttt{red}, \texttt{shape}(r') \Leftarrow \texttt{triangle})$.

We assume that any features unspecified by a causal function follow the principle of inertia,
and remain as they were before the causal interaction.

Note that although grammar $\mathcal{G}$ is very similar to a PCFG,
it is not context-free strictly speaking:
the ``bind feature'' production rule (Table~\ref{tab:core_grammar}, row 1) binds a feature to a lambda expression, and the subsequent steps within the scope of the $\lambda$-abstraction all refer to this feature.

For simplicity, we assume uniform transition probabilities for each production rule.
i.e., $\theta_l = \frac{1}{|I|}$ for each row $I$ with production rules $l\in I$.
By design, this grammar is inherently more likely to produce simpler expressions. The ``bind additional'' rule is called with probability 0.5,
and thus the number of conjunctions in the final expression follows a geometric decay with only 50\% combining more than one assertion, 25\% containing more than two, and so on.
The prior for a given expression is thus simply the product of all the productions that produced it:
\begin{equation}\label{equ:prior}
P_{\mathcal{G}}(f) = \prod_{l \in \Gamma}(\theta_{l})^{c_l}
\end{equation}
where $l \in I$ is the transition probability for production rule $l \in \Gamma$,
and $c_l$ is how many times rule $l$ was used for generating this causal function.

A causal function outputs result object(s) when particular agent and recipient objects are provided.
Take $\textsc{and}(\texttt{color}(r') \Leftarrow \texttt{color}(a), \texttt{shape}(r') \Leftarrow \texttt{square})$ for example.
For an agent $a$ that is a \texttt{red-circle} and
a recipient $r$ that is a \texttt{blue-pentagon}, $r$ will become $r'$: a \texttt{red-square}.
When a causal function $f$ involves a negation,
it could have produced more than one outcome.
For example let $w$ be
$\texttt{shape}(r') \Leftarrow \neg\texttt{triangle}$,
any object that is not triangular (and share the same color as $r$) is a possible option for being $r'$.
We further assume for simplicity that the different potential outcomes are equally probable,
and thus likelihood of a data point $d=(a, r, r')$ generated by a causal function $f$ is given by
\begin{equation}
P(d|f) = P(r'|f, a, r) =
  \begin{cases}
  \frac{1}{D(f(a,r))} &\text{if }r' \in D(f(a,r)),\\
  0 &\text{otherwise}
  \end{cases}
\end{equation}
where $D$ stands for \emph{domain} and $D(f(a,r))$ refers to the set of all possible result objects coming out of $f$ given agent $a$ and recipient $r$.
We initially assume a likelihood to 0 for any observation $a,r,r'$ incompatible of $f(a,r)$,
but later consider ``soft'' variants in which functional relationships are somewhat fallible.

\begin{table}[t]
  \caption{Example probabilistic grammar $\mathcal{G}$}
  \label{tab:core_grammar}
  \begin{tabular*}{\textwidth}{@{\extracolsep{\fill}} llll}
  \hline\noalign{\smallskip}
  Production rules     &            &                             &                             \\
  \noalign{\smallskip}\hline\noalign{\smallskip}
  Bind feature         & $S \to$    & $\lambda_{\phi_i}:A, \Phi$  &                             \\
  Bind additional      & $A \to$    & $B$                         & $\textsc{and}(B,S)$         \\
  Relation             & $B \to$    & $\phi_i(r') \Leftarrow C$   & $\phi_i(r') \Leftarrow \neg C$ \\
  Reference            & $C \to$    & $D$                         & $E$                         \\
  Relative reference   & $D$ $\to$  & $\phi_i(a)$                 & $\phi_i(r)$                 \\
  Absolute reference   & $E$ $\to$  & \texttt{value}$^{\phi_i}$   &                             \\
  \noalign{\smallskip}\hline
  \end{tabular*}
  Note:
  ``Bind feature'' samples a feature without replacement from the set of all features.
  $\phi_i$ in $D$ uses the feature selected in $A$,
  and value in $E$ is sampled uniformly from the support of the feature selected in $A$.
\end{table}

This framework naturally favors deterministic causal functions that are consistent with the evidence:
if a causal function predicts a specific result, when that outcome is indeed observed, likelihood will be 1.
In contrast, a causal function that predicts a range of outcomes will inevitably assign a lower likelihood to any one of these.

\section{Unpacking latent causal categories}\label{ap:model}

A CRP is a stochastic process widely used for creating partitions among entities \citep{aldous1985exchangeability}.
It draws on an analogy of sequentially seating infinite incoming customers to infinitely many tables in a Chinese restaurant,
where each table is also of infinite capacity.
The first observation $d^{(1)}$ is always assigned the first category $z^{(1)}$;
when $i > 1$, the probability for assigning category $z^{(i)}$ is given by
\begin{equation}
  \label{eq:crp}
  P(z^{(i)} = x|z^{(-i)}) =
  \begin{cases}
    \frac{\alpha}{i-1+\alpha}     &\text{if } x \text{ is a new category} \\
    \frac{|z^{(j)}|}{i-1+\alpha}  &\text{if } x = z^{(j)}
  \end{cases}
\end{equation}
where $z^{(j)}$ is an existing category,
and $|z^{(j)}|$ is the number of assigned objects in category $z^{(j)}$.
Parameter $\alpha$ ($\alpha > 0$) is known as the concentration, or dispersion parameter---the larger $\alpha$ is, the more likely a new object falls into a new category.
Holding the same $\alpha$,
categories with more members are preferred as they seem to be more ``common''.

Objects in a category characterize shared feature similarities,
modeled by a multinomial distribution over a finite number of feature values.
Let $\mu^{(z_i)} = [\mu_1, \ldots, \mu_n]$ be the mean feature vector of a given category $z_i$,
where each subscript $k$ in $\mu_k$ refers to a feature, $\mu_k$ is the mean value of feature $k$ for all the objects in category $z_i$,
the probability that an object is assigned to a particular category according to feature similarities is given by
\begin{equation}\label{equ:feats}
  P(o^{(i)}|\mu^{(z_i)}) = \prod_{k=1}^n \text{Bernoulli}(o^{(i)}; \mu_k)
\end{equation}
To compute $\mu_k$, let $o_v=[o_{v_1}, \ldots, o_{v_n}]$ be the feature values of an object $o$,
where each $v$ represents a feature value,
$o_{v_i}=1$ if object $o$ has this feature value and $o_{v_i}=0$ otherwise.
For a category $z = \{o^{(i)}, \ldots, o^{(m)}\}$,
$z_v := \sum_{j=1}^m o_v^{(j)}$,
which can be written as $z_v= [z_{v_1}, \ldots, z_{v_n}]$,
where $z_{v_i} = \sum_{j=1}^m o^{(j)}_{v_i}$.
Mean feature $\mu_k := \frac{z_{v_k}}{\sum_{l=1}^n(z_{v_l})}$.
We assign a Dirichlet prior to this multinomial distribution
in order to capture how important feature similarity is in forming categories.
Without leaning towards any specific feature,
the prior distribution over mean features is simply Dir($\beta$), $\beta \geq 0$.

It is not obvious whether mean features should be drawn from the agent object, recipient object, or both,
therefore we introduce one more hyper parameter $\gamma$, referring to the probability that mean feature is purely based on the agent:
when $\gamma=1$, categorization is only grounded on the agent objects,
when $\gamma=0$, only recipient's features are considered for categorization,
and when $\gamma=0.5$, both agent and recipient are considered equally.
We thus consider $0 \leq \gamma \leq 1$.

In sum, a Dirichlet Process that creates a distribution over causal category distributions according to Equation~\ref{equ:dp_core} has the priors:
\begin{align}\label{equ:dp_prior}
  z^{(i)}|z^{(-i)} &\sim \text{CRP}(\cdot|\alpha) \nonumber \\
  \mu^{(i)} &\sim \text{Dir}(\cdot|\beta) \nonumber \\
  f^{(z_i)} &\sim \mathcal{G}(\cdot)
\end{align}

And likelihoods are given by
\begin{align}\label{equ:dp_ll}
  a^{(i)}, r^{(i)}|\mu^{(z_i)} &\sim \text{Dir}(\ \cdot\ |\mu^{(z_i)}, \beta) \nonumber \\
  d^{(i)}|f^{(z_i)} &\sim f^{(z_i)}(a^{(i)}, r^{(i)})
\end{align}
where $\mu^{(z)}$ is the mean feature vector, and $f^{(z)}$ the assigned causal function.

To approximate the posterior with Gibbs sampling,
we construct a chain of samples where for each iteration, we sample a causal category for a random observation $d^{(i)}$ while fixing the category assignment to the other observations,
and a sampled causal category $z^{(i)}$ will then update the category parameters $\mu^{(z_i)}$ and $f^{(z_i)}$.
The category sampling step of this Gibbs sampler follows Equation~\ref{equ:dp_core},
and the local parameter update step follows definition of computing these parameters given objects in this category.
When the number of iterations $n \to \infty$, the sampled categories $\tilde{Z}_n$ coverges to the true posterior.

\section{Process variant}\label{ap:process}

\begin{algorithm*}[t]
  \caption{Process model}
  \label{algo:process}
  \begin{algorithmic}[1]
    \State Initialize an empty list of causal categories $Z$ \algorithmiccomment{Initialization}
    \State Assign $a^{(0)}, r^{(0)} \in d^{(0)}$ to category $z^{(1)}$, update $\mu^{(1)}$ \algorithmiccomment{Learning example goes to the first category}
    \State Sample $f^{(1)}$ from the learning posterior
    \State Record $z^{(1)}$ in list of causal categories $Z$
    \ForEach{$d^{(i)} \in D_G$}%
      \State sample $z^{(i)} \propto P(z^{(i)}|z^{(-i)})P(a^{(i)}, r^{(i)}|\mu^{(z_i)})$ \algorithmiccomment{Equation~\ref{eq:feat_cat}}
      \If{$z^{(i)} \in Z$} \algorithmiccomment{If current object belongs to an existing category}
        \State $r'^{(i)} \sim f^{(i)}(a^{(i)}, r^{(i)})$ \algorithmiccomment{Make prediction}
        \State Add $a^{(i)}, r^{(i)}$ to $z^{(i)}$: update $\mu^{(i)}$ \algorithmiccomment{Update $Z$}
      \Else
        \State Assign $a^{(i)}, r^{(i)}$ to a new category $z^{(k)}$: update $\mu^{(k)}$ \algorithmiccomment{Create a new category}
        \State Sample $f^{(k)}$ from the prior
        \State $r'^{(i)} \sim f^{(k)}(a^{(i)}, r^{(i)})$ \algorithmiccomment{Make prediction}
        \State Add $z^{(k)}$ to $Z$ \algorithmiccomment{Update $Z$}
      \EndIf
    \EndForEach
  \end{algorithmic}
\end{algorithm*}

The process model first assigns the object-pair in the learning example to an initial causal category $z^{(1)}$ governed by a causal law sampled from the posterior distribution $P(f|d)$. 
Crucially, for each generalization task,
it then assigns the encountered object pair scenario to either an existing causal category or a new category according to Equation~\ref{eq:feat_cat}.
If an existing causal category is selected, the model simply applies the requisite causal law category to make its prediction.
If a new category is sampled, however, a new causal law will be assigned to this category.
Since there is no evidence about what causal law may apply to this new category, this new causal law is sampled from the prior.
Algorithm~\ref{algo:process} shows this process.

Instead of approximating a posterior over infinitely many possible categories as the normative model, the process model maintains a small set of available categories that are created online as new generalizations are performed.
Furthermore, after categorizing an observation, the process model updates the list of causal categories $Z$ with this categorization decision, reflecting a commitment to its earlier decisions.
Hyperparameter $\alpha$ thus plays a slightly different role in the process model.
When $\alpha \to 0$, the process model becomes increasingly likely to stick with existing categories (Equation~\ref{eq:crp}).

\section{Experiment 1 setup}\label{ap:exp1-setup}

\begin{figure}[t]
  \centering
  \begin{subfigure}{.29\textwidth}
      \centering
      \includegraphics[width=\textwidth]{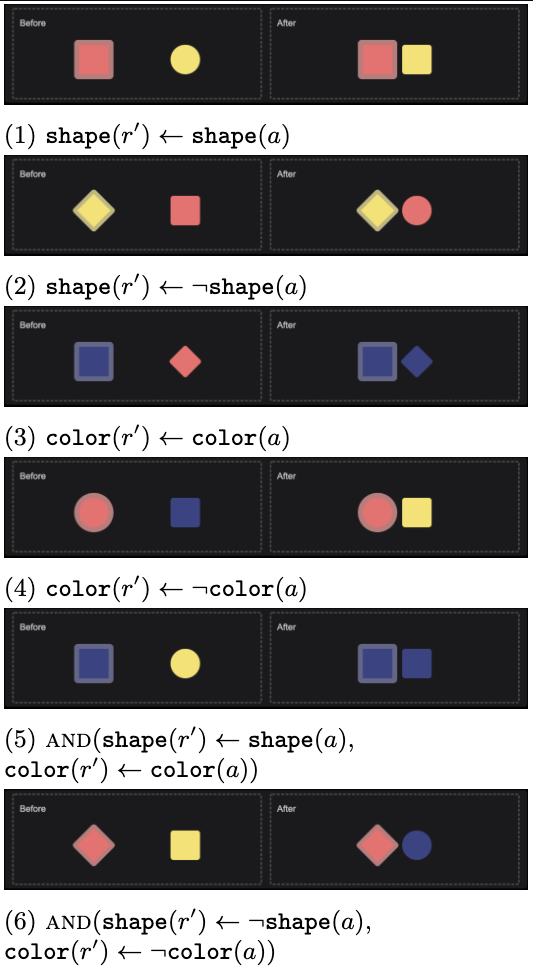}
      \caption{}
  \end{subfigure}
  \hfill
  \begin{subfigure}{.68\textwidth}
      \centering
      \includegraphics[width=\textwidth]{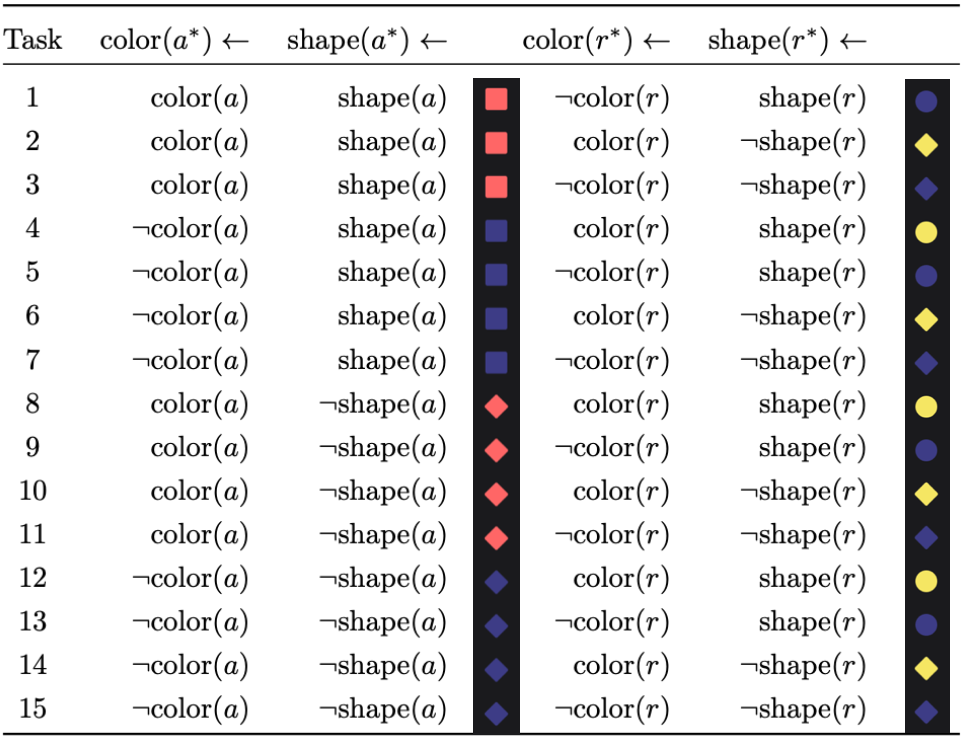}
      \caption{}
  \end{subfigure}
  \caption{Experiment 1 stimuli.
    (A): Learning conditions, showing objects before and after a causal interaction.
    (B): Generalization task configurations
    $a^*$, $r^*$ are the agent and recipient in each generalization task;
    $a$ and $r$ are the agents and recipients in the learning example.
    Example stones are for learning condition A1.}
  \label{fig:exp1}
\end{figure}

\subsection{Stimuli and design}

Participants were told that they were making predictions about the behavior of a magic world containing magic stones (agents) and normal stones (recipients).
In short videos, participants observed a  magic stone collide with a normal stone and appear to alter the normal stone's color and/or shape (see Figure~\ref{fig:exp_interface}). Magic stones had a thick border while normal stones had no border. We manipulated two object features---color \{\emph{red, yellow, blue}\} and shape  \{\emph{circle, square, diamond}\},
leading to $3 \times 3 = 9$ possible configurations for each object and a nominal $9 \times 9 \times 9 = 729$ configurations of agent and of recipient both pre- and post- the causal interaction.

We used a $6 \times 2$ between-subject design.
There were six learning examples varied between subjects (Figure~\ref{fig:exp1}A)---each participant saw one.
Each learning example demonstrates a causal effect differing in whether it results in a change to one or both features of the recipient object,
and whether either or both of these new values match the agent object's features.
Note that the function descriptions were not shown to participants and are by no means the only possible way to characterise the causal relationship being displayed.

For each learning example, we constructed 15 generalization tasks by varying object features systematically from the learning example (Figure~\ref{fig:exp1}B).
For example, A1 (see Figure~\ref{fig:exp1}A) depicts a \emph{red square} agent and a \emph{yellow circle} recipient,
and according to the specifications in Figure~\ref{fig:exp1}A,
task 1 for A1 has a \emph{red square} agent, and a \emph{blue circle} recipient.
We call the sequence of tasks from~1 to~15 ``near-first transfer'' because this sequence of tasks starts with those
that differ by only one feature from the learning example and progress to scenes in which all of the features differ.  Conversely, we call the sequence of tasks~15 to~1 the ``far-first transfer'' sequence,
because it starts with sets of stones that are completely different from those in the learning examples and progresses back to the more similar cases.
Within each sequence, whether the set of different-color tasks or the set of different-shape tasks appeared first (task 1 \& 2, 5 \& 6, 9 \& 10, 13 \& 14, 4---7 \& 8---11) was shuffled to counterbalance feature order.

\subsection{Procedure}

After instructions, participants had to pass a comprehension quiz to start the main task.
The main task contained a learning phase and a generalization phase.
During learning, participants watched one specific magic stone's effect on a normal stone (Figure~\ref{fig:exp_interface}A--C),
and they could replay the effect as many times as they wanted.
After that, participants were asked to make predictions for 15 new pairs of magic stones and normal stones sequentially,
by selecting from a panel of 9 possible stones (Figure~\ref{fig:exp_interface}D).
A summary of the learning example (as used in Figure~\ref{fig:exp1}A) was displayed at all times and the animation was replayed once between each generalization task to ensure it was not forgotten.
A demo of the task is available at \url{http://bramleylab.ppls.ed.ac.uk/experiments/bnz/magic_stones/index.html}.

\subsection{Model fits}\label{ap:exp1-fits}

Both the UnCala and LoCaLa models were fit to the behavioral data using the \texttt{optim} function in R.
As for the LoCaLaPro model, since it approximates posterior distribution with simulation-based method, we optimized parameter values via grid search.
Firstly, we set up a coarse grid with $\alpha$ = \{0.01, 0.1, 0.2, 0.3, 0.4, 0.5, 0.6, 0.7, 0.8, 0.9, 1, 1.5, 2, 4, 8\}, $\beta$ =\{0, 0.1, 0.2, 0.3, 0.4, 0.5, 0.6, 0.7, 0.8, 0.9, 1, 2, 4, 8, 16, 32, 64, 128, 256, 512, 1024\}.
After running this coarse grid and locating an optimal area, we ran another search over a finer grid for $\alpha=\{0.28, 0.30, 0.32, 0.34, 0.36, 0.38, 0.4, 0.42, 0.44, 0.46, 0.48, 0.5, 0.52\}$ ($\beta$ is the same as previously) to improve precision.

\section{Experiment 2 setup}\label{ap:exp2-setup}

\renewcommand\thesubfigure{\arabic{subfigure}}
\begin{figure}[t]
	\centering
  \begin{subfigure}[t]{0.47\textwidth}
  	\centering
  	\includegraphics[width=\linewidth]{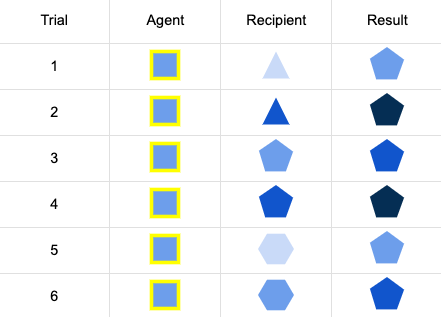}
  	\caption{Rule 1, fixed agent}
  \end{subfigure}
  \hfill
  \begin{subfigure}[t]{0.47\textwidth}
  	\centering
  	\includegraphics[width=\linewidth]{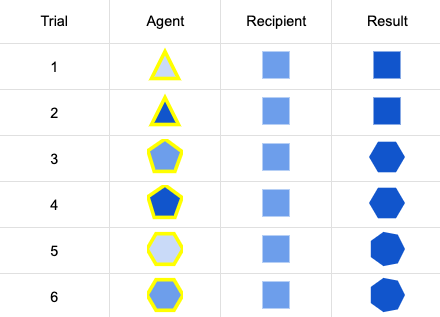}
  	\caption{Rule 1, fixed recipient}
  \end{subfigure}

  \begin{subfigure}[t]{0.47\textwidth}
    \centering
    \includegraphics[width=\linewidth]{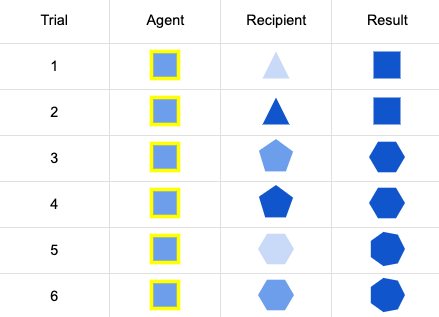}
    \caption{Rule 2, fixed agent}
  \end{subfigure}
  \hfill
  \begin{subfigure}[t]{0.47\textwidth}
  	\centering
  	\includegraphics[width=\linewidth]{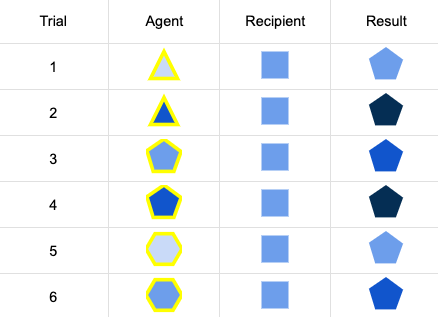}
  	\caption{Rule 2, fixed recipient}
  \end{subfigure}
\caption{Experiment 2 learning conditions.}
\label{fig:exp2-conds}
\end{figure}

\begin{table}[t]
  \renewcommand{\arraystretch}{1.5}
  \caption{Experiment 2 generalization task configurations}
  \begin{tabular*}{\textwidth}{@{\extracolsep{\fill}} llc|lc}
  \toprule
  & For the fixed object                        & Instance & For the varied object                  & Instance \\
  \midrule
  $o^* =$ & $\text{shade}(o), \text{edge}(o)$   & \adjustbox{valign=t}{\includegraphics[height=2em]{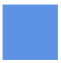}} & $\text{shade}(o), \neg \text{edge}(o)$ & \adjustbox{valign=t}{\includegraphics[height=2em]{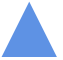}} \\
  & $\neg \text{shade}(o), \text{edge}(o)$      & \adjustbox{valign=t}{\includegraphics[height=2em]{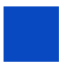}} &                                        & \adjustbox{valign=t}{\includegraphics[height=2em]{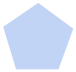}} \\
  & $\text{shade}(o), \neg \text{edge}(o)$      & \adjustbox{valign=t}{\includegraphics[height=2em]{s23}} & $\neg \text{shade}(o), \text{edge}(o)$ & \adjustbox{valign=t}{\includegraphics[height=2em]{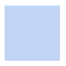}} \\
  & $\neg \text{shade}(o), \neg \text{edge}(o)$ & \adjustbox{valign=t}{\includegraphics[height=2em]{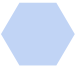}} &                                        & \adjustbox{valign=t}{\includegraphics[height=2em]{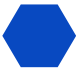}} \\
  \bottomrule
  \end{tabular*}
  $o^*$ is the object in generalization tasks, $o$ is the object shown during learning.
  For the varied object, $\neg \text{shade}(o)$ means picking a shade that has not appeared during the learning phase, and we chose two instances for it.
  \label{tab:exp2-gs}
\end{table}

\subsection{Stimuli and design}

Similar to Experiment 1, we varied the shape and color properties of the objects.
However, instead of using categorical values,
we introduced intuitively ordinal feature values.
Shapes were all equilateral and differed in terms of their number of sides:
3 (triangle),
4 (square),
5 (pentagon),
6 (hexagon), and
7 (heptagon);
colors were of identical hue and saturation (blue) but differed in lightness varying between:
1 (light blue \texttt{\#c9daf8}),
2 (medium blue \texttt{\#6d9eeb}),
3 (dark blue \texttt{\#1155cc}), and
4 (very dark blue \texttt{\#052e54}).
Staying within the features' observed values this leads to $4 \times 4 = 16$ possible configurations for each object,
and a nominal $16^3 = 4096$ possible configurations for objects both pre- and post- the causal interaction.
These ordinal features enlarge the space of effects and greatly enriches the space of plausible rules, for example
allowing causal laws in which a recipient stone becomes \emph{darker} or \emph{lighter} when acted upon, gaining or losing sides, as well as those involving copying or taking specific or random values.

During learning, each participant observed six causal interactions between different pairs of agent and recipient before making generalizations.
We included 2 (evidence-balance) $\times$ 2 (ground truth) between-subject factors (see Figure~\ref{fig:exp2-conds}).
with respect to evidence-balance,  for \emph{fixed-agent} conditions B1 and B3, an identical agent was shown in all learning examples, while the recipients it acted on were varied systematically;
in the \emph{fixed-recipient} conditions B2 and B4, the recipient object was always identical but was acted on by six different agents.
We designed the evidence to be consistent with two ``ground truth'' rules that counterbalance between the roles of the shape and the color features:
\begin{enumerate}[align=left]
  \item [\textbf{Rule 1}] (B1/B2) The recipient gets one increment darker and takes the agent's shape plus one edge \\ $\textsc{AND}(\texttt{edge}(r')\leftarrow\texttt{edge}(a)+1, \texttt{shade}(r')\leftarrow\texttt{shade}(r)+1)$
  \item [\textbf{Rule 2}] (B3/B4) The recipient gains an edge and takes the agent's shade plus one shade increment \\ $\textsc{AND}(\texttt{shade}(r')\leftarrow\texttt{shade}(a)+1, \texttt{edge}(r')\leftarrow\texttt{edge}(r)+1)$
\end{enumerate}
Note that these ``ground truth'' rules are just one of an unbounded set of possible universal causal relations consistent with the six learning trials, and a single universal category is just one of a much larger set again of possible local causal law category structures. 

We composed generalization tasks according to the configurations in Table~\ref{tab:exp2-gs}.
In total there were $4 \times 4 = 16$ generalization tasks for each condition.

\subsection{Procedure}

After completing instructions, participants had to pass a comprehension quiz to proceed to the main task, consisting of a learning phase, self-report, and a generalization phase.
After the main task, participants provided demographic information and feedback.
A demo of the task is available at \url{http://bramleylab.ppls.ed.ac.uk/experiments/bnz/myst/p/welcome.html}.

Each participant was randomly assigned to one of the four learning conditions (Figure~\ref{fig:exp2-conds}).
The six pairs of agent and recipient stones were shown in random order, one after another.
By clicking a ``Test'' button, they could watch the causal interaction as many times as they wanted.
After each object pair was tested, a summary visualization of the agent, recipient and the result was added to the top of the page (see Figure~\ref{fig:exp_interface}E--F), and remained visible for the rest of the task.
After the learning phase, participants were asked to write down their best guess about how the mysterious stones worked, and told they would receive a \$0.50 bonus if they described the true underlying causal law.
In the generalization phase, participants faced the 18 generalization trials sequentially in random order.
For each, participants predicted the result recipient by selecting
a number of edges and the shade of blue from two drop-down menus (see Figure~\ref{fig:exp_interface}F).
Participants were instructed they would receive a \$0.10 for each correct prediction.
We bonused participants as described afterwards.

\subsection{Generalization consistency}\label{ap:exp2-gen}

\begin{figure*}[t]
  \centering
  \includegraphics[width=\textwidth]{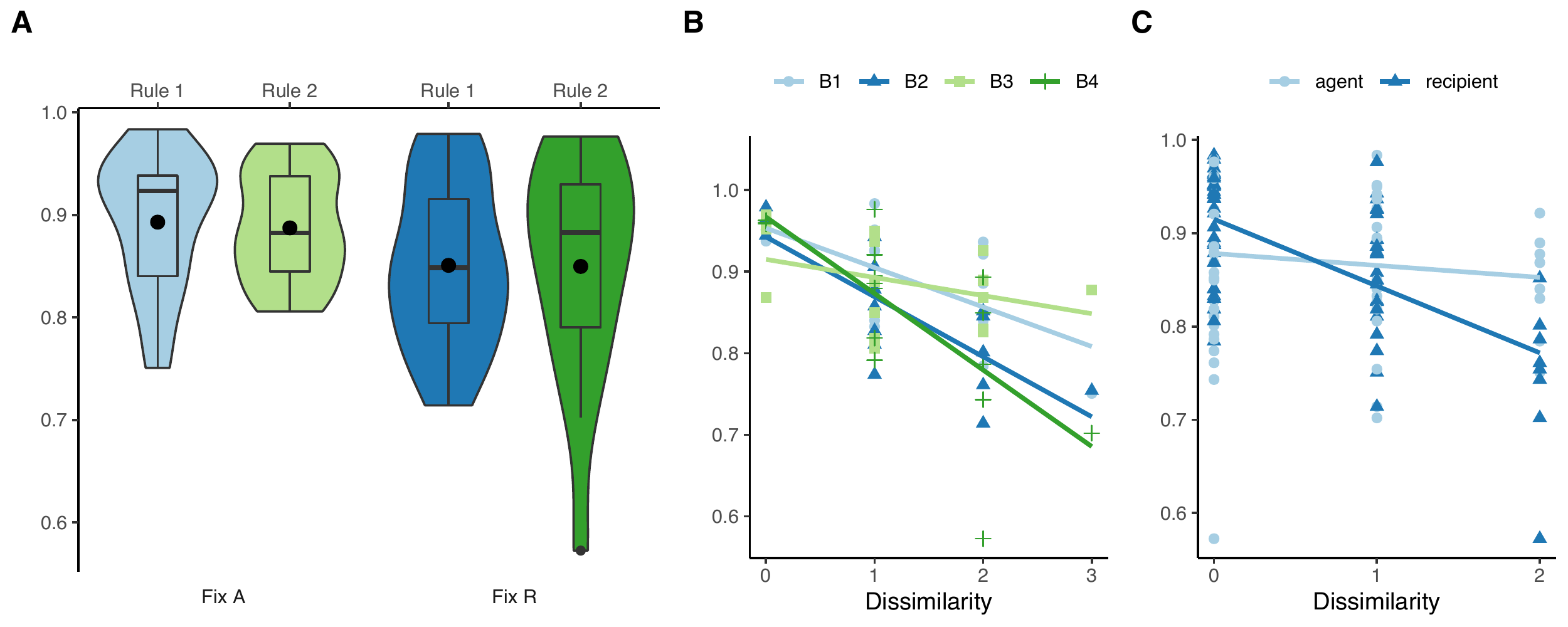}
  \caption{Behavioral results of Experiment 2.
  All $y$-axes are Cronbach's alpha values.
  A. Task-wise inter-person consistency per condition.
  Violin plots are density. Black dots are mean Cronbach's alpha values per condition. The major bar in the box plot is median and box extent is the 25 and 75 quantiles.
  B. Inter-person consistency per task differences.
  C. Inter-person consistency per role differences.
  }
  \label{fig:myst-gendata}
\end{figure*}

As with Experiment 1, we measured inter-person consistency in generalization predictions computing $\rho_T$ for the sixteen generalization tasks per condition (excluding the two catch-trials), totalling $4 \times 16 = 64$ values.
Mean consistency was $\rho_T = 0.87 \pm 0.08$, with min $\rho_T = 0.57$, max $\rho_T = 0.98$.
To compare generalization consistency against random selections, for each condition we conducted Fisher's exact test on the contingency table of selecting each possible result per trial.
For all four conditions, $p < 0.001$.
Thus, as in Experiment 1, participants produced systematic generalization patterns.

We then compared inter-person generalization consistency by condition.
As illustrated in Figure~\ref{fig:myst-gendata}A, the {\em fixed-agent} condition induced higher consistency ($\rho_T = 0.89 \pm 0.06$) than the {\em fixed-recipient} condition ($\rho_T = 0.85 \pm 0.1$), $t(31) = 2.12, p = 0.04, 95\%\text{CI} = [0.001, 0.08]$,
while the difference in $\rho_T$ between the ground truth condition was negligible, $t(31) = 0.22, p=\text{n.s.}$. No interaction was detected. 
In short, participants made more homogeneous predictions after observing the same agent acting on a range of recipients,
and diverged more having observed different agents interacting on the same recipient.

Generalization consistency decreased as objects in the generalization tasks become more distinct from those in the learning examples (Figure~\ref{fig:myst-gendata}B). 
To show this, we constructed a rough measure of {\em dissimilarity}, by counting the features of generalization trials that took novel values never observed in the learning phase.
Formally, let $F_L$ be the set of unique feature values of all the objects appeared during learning, and $F_i$ be the set of unique feature values of objects in a generalization trial $i$, dissimilarity score $DS = |F_i \setminus F_L|$.
By design, dissimilarity scores $DS \in \{0,1,2,3\}$ (Table~\ref{tab:exp2-gs}).
We found a significant negative relationship between task dissimilarity and generalization consistency, $\beta = -0.06, F(1, 62) = 37.48, p < 0.001$.

Finally, we fit a linear regression model predicting $\rho_T$ with task dissimilarity, evidence-balance and ground truth, $F(3, 60) = 15.63, p < 0.001$.
This revealed main effects of dissimilarity ($\beta = -0.06, p < 0.001$) and evidence-balance ({\em fixed-recipient}, $\beta = -0.04, p = 0.01$), but not ground truth ({\em rule 2}, $\beta = -0.003, p = \text{n.s.}$). 
As depicted in Figure~\ref{fig:myst-gendata}B, consistency of judgments in the fixed-agent conditions (B1 \& B3, lighter lines) decreased slower than the fixed-recipient conditions as dissimilarity increased (B2 \& B4, darker lines).

Not only did the evidence-balance condition have a significant effect on generalization consistency, dissimilarity of the agent or recipient objects in the generalization tasks was also associated with lower consistency (Figure~\ref{fig:myst-gendata}C).
Holding recipient dissimilarity constant,
increasing agent dissimilarity does not predict prediction consistency significantly,
$F(1, 62) = 0.77, p = \text{n.s.}$;
however, recipient dissimilarity does,
$F(1, 62) = 38.8, p <0.001$.

\subsection{Self-reports}\label{ap:exp2-selfr}

\begin{figure*}[t]
  \centering
  \includegraphics[width=\textwidth]{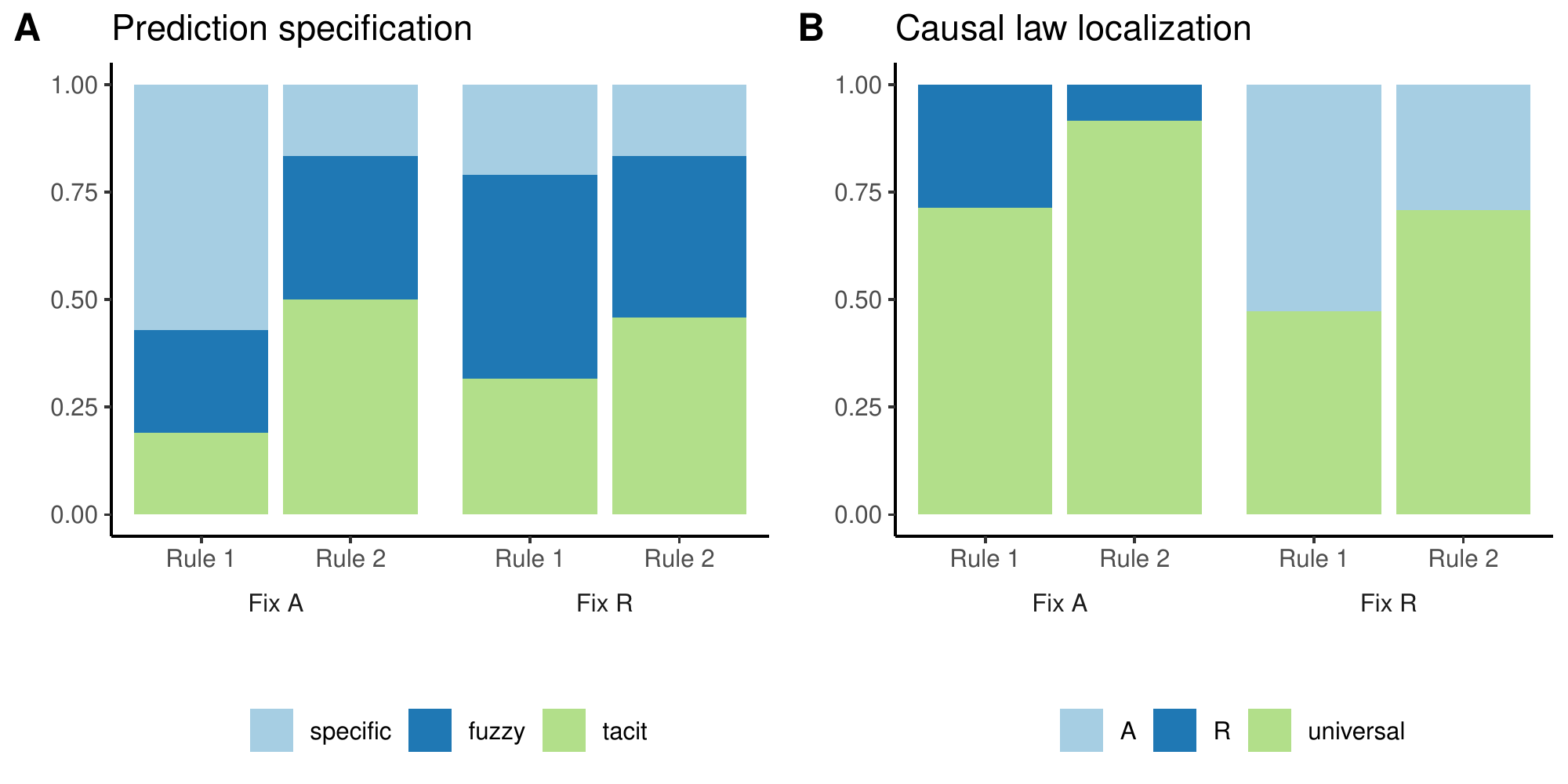}
  \caption{Rule guess categories.}
  \label{fig:gen-labels}
\end{figure*}

In Experiment 2, we asked participants to provide an explicit free-text guess about the nature of the causal relationship(s) being tested after they completed the learning phase.
Eighty-six percent of these total responses (88/102) were compatible with the relevant learning observations,
and here we only analyze these.
Two independent coders categorized participants guesses according to their specificity and implicit localization of causal powers.
The first coder categorized all free responses, and the categorized 15\% were then compared against the first coder's. Agreement level was 92\%.
The full set of free responses and the detailed coding scheme are available at \url{https://github.com/bramleyccslab/causal_objects}.

Since our ground truths are not the only rules consistent with the learning data, we analyzed participant self-reports not according to whether they got the ground truths right, but whether their own rules were consistent with the learning data, as well as the level of generality in the reports.
Hence, we first defined three exclusive and exhaustive response specificity categories: {\em specific}, {\em fuzzy}, and {\em tacit}. A {\em specific} self-report would predict a unique result object for any potential combination of agent and recipient (for example ``The inactive shape is always changed to a pentagon \& its shade is changed to one step darker than the active stone''). A {\em fuzzy} rule was one that left open for more than one possible result objects (for example ``It will be different colors and shapes'').
We distinguished a second form of under-specified self-report, {\em tacit}, if it left a feature unmentioned, which depending on background assumptions might be taken to imply that feature remained unchanged but could also be compatible with it taking some new or random value (for example ``The active stone adds a side to the inactive stone'').

We also had the coders categorize responses according to whether and how a self-report localized the domain of the causal law asserted.
Concretely, we included four labels {\em A, R, AR}, and {\em universal}. 
If a response mentioned a specific context of influence, typically using an \emph{if...} clause, we labelled this according to whether the context mentioned the Agent (e.g. ``If the active stone is darker than the inactive stone, it turns the inactive stone darker''), Recipient (e.g. ``The active stone causes the other stones to change into a pentagon shape, unless it is already a pentagon shape, in which case it makes it darker''), or both.
If a response made no localization or context (e.g. ``The active stone cause inactive stones to five sided stone'') then it was labeled as {\em universal}.

Figure~\ref{fig:gen-labels} illustrates the coding results by learning condition.
Guess specificity is summarized in Figure~\ref{fig:gen-labels}A.
We fit a multinomial logistic regression model predicting specificity by evidence-balance and ground truth factors,
and found that when taking the {\em specific} self-report type as baseline, the ground truth factor is a significant predictor for the {\em tacit} type ($\beta = 1.54, p = 0.008$), while evidence-balance is not ($\beta = 0.78, p =\text{n.s.}$).
Neither of these two factors are significant for the {\em fuzzy} type.
Figure~\ref{fig:gen-labels}B summarizes participants' guesses in terms of localization.
No participant localized their rule in terms of both Agent and Recipient.
Unsurprisingly, whenever localization occurred, it was applied with respect to the object that varied during the learning phase.
A logistic regression predicting universal rule probability by condition showed that both evidence-balance ({\em fixed-recipient}, $\beta = -1.21, z = -2.3, p = 0.02$) and ground truth ({\em rule 2}, $\beta = 1.17, z = 2.3, p = 0.02$) were associated with more universal rules.
There was no evidence for an interaction, $z = -0.5, p=\text{n.s.}$.

\subsection{Model fits}\label{ap:exp2-fits}

We extended the grammar used in Experiment 1 to cover a larger space of ordered feature relationships.
Concretely, we introduced \textsc{+1, -1, >, <} at the ``bind relation'' step to accommodate potential assertions about the ordering of feature values used in this experiment.
As in Experiment 1, LoCaLa was expensive to evaluate so we optimised its parameters using a coarse grid search.
Since there were six data points, during each iteration of the Gibss sampler, when $\alpha = 5$ this observation has a half-half chance to create a new causal category or join the rest, in terms of category size preference, and this chance grows as $\alpha$ increases (Equation~\ref{eq:crp}). Therefore, we centered the support values for $\alpha$ round $5$, with an exponential increase for larger values, resulting in $\alpha \in \{1,2,3,4,5,6,7,8,9,10,16,32,64,128,256\}$.
$\beta$ takes the same range of values as in fitting the models in Experiment 1.
For $\gamma$, values of $\gamma = 1, 0.5$ and $0$ are of particular theoretical interest,
representing localization based on just the agent, agent and recipient equally, or just the recipient. We also included $\gamma = 0.25$ and $\gamma = 0.75$ consistent with a mixed focus biased toward either agent or recipient.

\end{document}